\documentclass{article}
\usepackage[preprint]{colm2026_conference}
\usepackage{microtype}

\usepackage{amsmath,amsfonts,bm}

\def\eqref#1{equation~\ref{#1}}

\def\1{\bm{1}}

\DeclareMathAlphabet{\mathsfit}{\encodingdefault}{\sfdefault}{m}{sl}
\SetMathAlphabet{\mathsfit}{bold}{\encodingdefault}{\sfdefault}{bx}{n}

\usepackage{float}
\usepackage{flafter}
\usepackage{placeins}
\usepackage{hyperref}
\definecolor{darkblue}{rgb}{0, 0, 0.5}
\hypersetup{colorlinks=true,linkcolor=darkblue,citecolor=darkblue,urlcolor=darkblue}
\usepackage{url}
\usepackage{booktabs}
\usepackage{amsmath}
\usepackage{amssymb}
\usepackage{multirow}
\usepackage{xcolor}
\usepackage{colortbl}
\definecolor{medalgold}{HTML}{F7DC8A}
\definecolor{medalsilver}{HTML}{DADDE2}
\definecolor{medalbronze}{HTML}{EEC7A2}
\newcommand{\gold}[1]{\cellcolor{medalgold}#1}
\newcommand{\silver}[1]{\cellcolor{medalsilver}#1}
\newcommand{\bronze}[1]{\cellcolor{medalbronze}#1}
\usepackage{xspace} 
\usepackage{enumitem}
\usepackage{pifont} 
\usepackage{tikz}
\usepackage{pgfplots}
\usetikzlibrary{positioning,backgrounds,fit,arrows.meta,matrix}

\pgfplotsset{compat=1.18}

\newcommand{\ci}[2]{{\scriptsize [#1,\,#2]}}
\newcommand{\wq}{\mathrm{PA}}

\newcommand{\method}{\textsc{RINSE}\xspace}

\newif\ifcomment
\commentfalse

\ifcomment
    \newcounter{YXNumberOfComments}
    \stepcounter{YXNumberOfComments}
    \DeclareRobustCommand{\xym}[1]{\textcolor{orange}{\small \bf [XYM\#\arabic{YXNumberOfComments}\stepcounter{YXNumberOfComments}: #1]}}

    \newcounter{PHNumberOfComments}
    \stepcounter{PHNumberOfComments}
    \DeclareRobustCommand{\hpc}[1]{\textcolor{blue}{\small \bf [HPC\#\arabic{PHNumberOfComments}\stepcounter{PHNumberOfComments}: #1]}}

    \DeclareRobustCommand{\del}[1]{{{\color{red}\st{#1}}}}
\else
    \DeclareRobustCommand\xym[1]{}
    \DeclareRobustCommand\hpc[1]{}
    \DeclareRobustCommand{\del}[1]{}
\fi

\title{Relevance Is Not Sufficient Evidence: Detecting Evidence Gaps Before Generation in RAG}

\author{Suting Chen$^{*1}$ \quad Peichun Hua$^{*2}$ \quad Yunming Xiao$^{2\dagger}$ \\
$^{1}$Northwestern University \quad $^{2}$The Chinese University of Hong Kong, Shenzhen \\
\texttt{suting.chen@u.northwestern.edu} \\
\texttt{peichunhua@link.cuhk.edu.cn} \quad \texttt{yunmingxiao@cuhk.edu.cn}
}

\begin{document}

\maketitle
{\renewcommand{\thefootnote}{\fnsymbol{footnote}}\footnotetext[1]{Equal contribution.}\footnotetext[2]{Corresponding author.}}

\begin{abstract}
Retrieval-augmented generation (RAG) grounds large language models (LLMs) in external sources. However, retrieved passages often name the right entities while lacking the actual facts an answer requires.
Generators rarely notice: even with an explicit instruction to abstain, 12 generators still answer 40.0--99.3\% of insufficient-evidence questions. A common remedy trains abstention into the generator through fine-tuning or reinforcement learning. This ties the decision to one model's weights, potentially rewarding correct answers recalled from its parametric knowledge as if the evidence had supported them. The decision arrives only as the generator's own output, so every insufficient-evidence question still costs a full generator call.
We ask instead whether sufficiency can be judged from the question and evidence alone, \textit{before} any answer exists.
We first demonstrate several pitfalls in constructing tests for insufficient evidence. Directly removing the relevant evidence or pairing evidence with unrelated questions can introduce \textit{cues} that reveal the label without testing sufficiency, such as lexical overlap and evidence position.
We therefore build a paired benchmark from substitution, deletion, and question-swap constructions that vary answer support while controlling selected surface features, such as word use.
On this benchmark, sufficiency can be judged without generating an answer, but no single signal works on every dataset. We introduce \method (\textbf{R}elevance \textbf{I}s \textbf{N}ot \textbf{S}ufficient \textbf{E}vidence), which combines three signals, one for each way evidence falls short: whether every part of the question is covered, whether any passage actually offers an answer, and whether a small language model, reading the passages together, treats them as sufficient.
Across six datasets, \method ranks sufficient above insufficient evidence with a score of 0.837 (chance 0.5), ahead of the best of 10 prior methods (0.746) and of a frontier model asked the same question through an API (0.784). Its weakest dataset scores higher than any other method's weakest (0.684 vs. 0.676). It runs locally before generation, taking only 36.5 ms per question on a single GPU.
\end{abstract}

\section{Introduction}

A large language model (LLM) answers based on its parameters, which are fixed at training time and do not hold private information due to proprietary concerns \citep{carlini2021extracting}, or more recent information due to the high cost of continual learning \citep{jang2022continual}. Asked beyond what they store, models often hallucinate \citep{huang2025hallucination}, producing a fluent and confident answer that no source supports \citep{mallen2023popqa}. Retrieval-augmented generation (RAG) was introduced to provide external sources that are more relevant and timely \citep{lewis2020rag}. However, relevant passages are not always sufficient: a passage can name the right people, places, and events but miss the one fact or connection that the question depends on.

Whether the passages suffice decides what a RAG system does next: answer, retrieve again, widen the search to the live web, or abstain \citep{yan2024crag,moskvoretskii2025adaptive}. Retrieval itself gives no signal for this choice. A web cache reports a miss when a requested page is absent, but a semantic search always returns its top-$k$ passages, even when the answer lies outside its index. The Semantics Delivery Network (SemDN), a shared retrieval layer that serves passages to LLM agents on behalf of websites, therefore needs a coverage signal to decide when to fetch new content \citep{hua2026semantics}. Without one, it either returns incomplete evidence as if it were complete or repeats external search and website fetches on every query. Agentic search makes the same choice at every turn, between answering and retrieving more \citep{jiang2023flare}. Because the check runs on every query, it must cost far less than the generation it saves. This raises the question: \textbf{how can a RAG system judge whether the retrieved passages contain enough information to answer the question?}


A RAG system cannot rely on its generator to notice the gap. Given insufficient context, models often answer from possibly stale parametric knowledge instead of abstaining \citep{joren2025sufficient,peng2025unanswerability}. \citet{joren2025sufficient} reports that the wrong-answer rate rises from $10.2\%$ without context to $66.1\%$ with insufficient context. We instruct 12 generators from 7 families (135M--32B parameters) to reply \emph{unanswerable} when the documents lack the answer, on 300 sufficient and 300 insufficient records from the six evaluation datasets. They still answer $40.0\%$ to $99.3\%$ of the insufficient records, and they also refuse up to $29.7\%$ of the sufficient ones (Table~\ref{tab:motivation}). Each such answer costs a full generation: 7--32B models take a median of 150--855~ms for an answer of up to 24 tokens. An evidence gap should therefore be caught before the generator is called.

\begin{table}[t]
\caption{Twelve generators instructed to reply \emph{unanswerable} when the documents lack the answer, on 300 sufficient and 300 insufficient records drawn evenly from the six datasets of Table~\ref{tab:grid}. \emph{Answers}: share of insufficient records answered; \emph{Refuses}: share of sufficient records refused; \emph{ms}: median latency of a greedy answer of up to 24 tokens, with passages capped at 1{,}600 characters.}
\label{tab:motivation}
\centering\footnotesize
\setlength{\tabcolsep}{3.4pt}
\begin{tabular}{@{}lrrr@{\hspace{1.5em}}lrrr@{}}
\toprule
\textbf{Model} & \textbf{Answers (\%)} & \textbf{Refuses (\%)} & \textbf{ms} &
\textbf{Model} & \textbf{Answers (\%)} & \textbf{Refuses (\%)} & \textbf{ms} \\
\midrule
Qwen3-0.6B & 53.7 & 29.7 & 131 & SmolLM2-135M & 99.3 & 0.3 & 358 \\
Qwen3-1.7B & 64.3 & 13.3 & 147 & Llama-3.2-1B & 81.7 & 12.7 & 92 \\
Qwen3-4B & 55.3 & 16.0 & 211 & Mistral-7B-v0.2 & 68.0 & 7.7 & 448 \\
Qwen3-8B & 61.3 & 11.3 & 205 & Falcon3-7B & 68.0 & 16.3 & 150 \\
Qwen3-14B & 40.7 & 27.0 & 331 & Zephyr-7B-beta & 82.7 & 3.0 & 469 \\
Qwen3-32B & 48.7 & 11.7 & 855 & Phi-4 & 40.0 & 18.3 & 588 \\
\bottomrule
\end{tabular}
\end{table}


A natural response is to train the generator to abstain through supervised fine-tuning \citep{zhang2024rtuning}, preference alignment \citep{song2025trustalign}, or reinforcement learning with rewards that separate correct answers, hallucinations, and abstentions \citep{wei2025truthrl,zhao2026grace}.
However, because the supervision comes from answer outcomes, what such training teaches is the \textit{generator's own knowledge boundary} rather than a \textit{judgment about the evidence}: the model is rewarded for answering the questions it knows and for abstaining on those it does not, regardless of which passages sit in the prompt.
The two coincide only when the model's knowledge already matches the retrieved evidence, which is the assumption RAG exists to remove. On NeoQA~\citep{glockner2025neoqa}, built from fabricated news so that answers must come from the supplied passages, the four methods that read a generated answer rank the sufficient evidence set above the insufficient one barely above chance ($0.503$--$0.520$, where chance is $0.5$). Our method, which reads only the question and the evidence, reaches $0.773$, and its content-coverage reader alone reaches $0.799$ (Section~\ref{sec:autorater}). In addition, such training-based methods are bound to the model being trained, so they must be repeated for every model update, cannot be installed on a closed-source model exposed only through an API, where a user supplies training data but not the training objective, and can erode under later training with other objectives \citep{song2025hallucinationtax,kirichenko2025abstentionbench}. We argue, instead, that \textbf{whether a set of passages answers a question is a property of the question and the retrieved passages}, not of the model that will read them. We therefore ask \textbf{whether that property can be computed cheaply and without generating an answer}.

Given a question $q$ and a set of passages $E$, the task is to decide whether $E$ contains the information needed to derive a complete answer, including any intermediate facts, before generation begins.
Related methods answer neighboring questions: relevance rankers ask what a passage is about \citep{sachan2022upr,es2024ragas}, verification and confidence methods read an answer once it exists \citep{honovich2022true,kadavath2022ptrue}, and retrieval controllers ask what a particular model still needs \citep{cheng2024uar}. The closest existing work is Sufficient Context \citep{joren2025sufficient}, which judges the same property with a large autorater---a frontier model at best, a fine-tuned 11B entailment model at its smallest; whether a \textit{lightweight} local scorer can make the same decision remains open.

A benchmark for this task compares passages that support an answer with passages that do not, so anything else that differs between the two sides becomes a \textit{cue} a scorer can use in place of sufficiency. For example, the share of the question's words that appear in the passage separates answerable from unrelated questions on SQuAD~2.0 almost perfectly ($0.993$, a \textit{lexical cue}), yet falls to $0.573$, near chance, once the unanswerable question is written about the passage.

We build a paired benchmark whose sufficient and insufficient sides share a question or an evidence set, and we audit the remaining construction cues.
We use three constructions: 1) keep the question fixed and swap a needed passage for a related distractor; 2) pair one passage with questions written to be about it but unanswerable from it; 3) delete a needed passage from evidence sets whose sizes already vary widely, so that passage count reveals little about the label.

These constructions expose three different ways that evidence falls short. It can leave out something that the question asks for, stay on topic while offering no answer at all, or carry the answer across several passages, none of which is sufficient alone \citep{ferguson2020iirc,trivedi2022musique}. Inspired by these failure modes, we propose \method, a lightweight scorer with one reader for each, checking whether every part of the question is covered, whether any passage offers an answer, and whether a small language model, reading the passages together, treats them as sufficient. Each signal is corrected for the size of the evidence set, and the three are summed with equal weights.

Averaged over six datasets, \method\ scores $0.837$, ahead of all 10 prior methods, including Self-RAG~\citep{asai2024selfrag} at $0.746$ and UPR~\citep{sachan2022upr} at $0.721$. It is also the most consistent: its weakest dataset scores $0.684$, while every other local method drops to $0.644$ or lower on at least one. Asked directly whether the evidence suffices, a local 1.5B model scores $0.540$ and a frontier model queried through an API scores $0.784$; \method\ ranks above both. On evidence returned by real retrievers, \method\ scores $0.813$, against $0.674$ for the strongest prior method. \method\ needs neither a generated answer nor an API call: three lightweight components, the largest of them 1.5B parameters, take 36.5~ms per record on a single GPU (Section~\ref{sec:autorater}), less than the median time of any of the 12 short generations in Table~\ref{tab:motivation} (92--855~ms).


\section{Related Work}
\label{sec:setting}

The closest line of work judges \textbf{evidence sufficiency} directly. Sufficient Context \citep{joren2025sufficient} asks whether a diligent reader could answer from the provided context alone, the same target as \method, and labels it with an autorater ranging from an 11B entailment model to a frontier LLM; S2G-RAG \citep{li2026s2g} makes the same judgment inside an iterative retrieval loop. We compare with Sufficient Context on both a local and a frontier model, and study how negative construction shapes the evaluation. A second line performs \textbf{relevance scoring}: UPR \citep{sachan2022upr}, RAGAS \citep{es2024ragas} and CRAG \citep{yan2024crag} judge whether a passage is about the question, yet a relevant passage can omit the fact the answer needs, and a per-passage grade misses support spread across passages. Many methods instead read \textbf{generator-side signals}. Retrieval controllers such as UAR \citep{cheng2024uar}, FLARE \citep{jiang2023flare} and Self-RAG \citep{asai2024selfrag}, and answer-side confidence estimates such as P(True) \citep{kadavath2022ptrue}, semantic entropy \citep{kuhn2023semantic,farquhar2024semantic} and attribution \citep{honovich2022true,gao2023alce}, all depend on one generator's hidden states or outputs, whereas \method\ reads only the question and the evidence. Methods for \textbf{trained abstention}, including R-Tuning, Trust-Align, TruthRL and GRACE \citep{zhang2024rtuning,song2025trustalign,wei2025truthrl,zhao2026grace}, teach one generator to refuse, and later fine-tuning erodes that behavior \citep{song2025hallucinationtax,kirichenko2025abstentionbench}; a generator-independent score could supply their rewards with a sufficiency signal. Finally, work on \textbf{constructed negatives} found hypothesis-only and lexical-overlap artifacts in natural language inference \citep{gururangan2018artifacts,poliak2018hypothesis,mccoy2019right}, and SQuAD~2.0 replaced generated unanswerable questions with authored near-misses \citep{rajpurkar2018squad2}; our cue audit applies the same scrutiny to evidence sufficiency (Section~\ref{sec:protocol}). Appendix~\ref{app:related} gives details.

\section{Constructing and Auditing Sufficiency Contrasts}
\label{sec:protocol}

\begin{figure}[t]
\centering
\definecolor{ovInk}{HTML}{344151}
\definecolor{ovFit}{HTML}{2E7D73}
\definecolor{ovChange}{HTML}{B8672E}
\resizebox{\linewidth}{!}{%
\begin{tikzpicture}[x=1cm,y=1cm,
  pool/.style={draw=ovInk!45, fill=white, rounded corners=2pt,
    line width=.5pt, inner sep=3pt, align=left, font=\footnotesize,
    minimum height=1.18cm},
  support/.style={pool, draw=ovFit!75, fill=ovFit!12},
  distractor/.style={pool, draw=ovChange!75, fill=ovChange!12},
  changed/.style={draw=ovChange!85, fill=ovChange!7,
    rounded corners=2pt, line width=.6pt, inner sep=3pt,
    align=left, font=\footnotesize},
  cell/.style={anchor=west, align=left, font=\footnotesize, text=ovInk},
  head/.style={anchor=west, font=\scriptsize\bfseries, text=ovInk!80},
]
\path[use as bounding box] (0,0.08) rectangle (14.9,-4.86);

\node[pool, text width=3.55cm, anchor=north west] at (0,0.04)
  {\textbf{$q$} \emph{When} was \emph{Parasite}'s director born?};
\node[support, text width=3.25cm, anchor=north west] at (3.94,0.04)
  {\textbf{$d_1$} \emph{Parasite} was directed by Bong Joon-ho.};
\node[support, text width=3.25cm, anchor=north west] at (7.58,0.04)
  {\textbf{$d_2$} Bong Joon-ho was born in 1969.};
\node[distractor, text width=3.45cm, anchor=north west] at (11.22,0.04)
  {\textbf{$d_3$} \emph{Parasite} won an award in 2019.};

\begin{scope}[yshift=0.28cm]
\node[head] at (0.12,-1.73) {PAIR};
\node[head] at (2.63,-1.73) {QUESTION};
\node[head] at (5.95,-1.73) {SELECTED PASSAGES};
\node[head] at (9.05,-1.73) {SUFFICIENCY};
\node[head] at (11.40,-1.73) {CUE ANALYSIS};
\draw[ovInk!55, line width=.55pt] (0,-1.94) -- (14.9,-1.94);

\fill[ovFit!6] (0,-1.95) rectangle (14.9,-2.73);
\node[cell, text width=2.27cm] at (0.12,-2.34) {\textbf{Reference}};
\node[cell, text width=2.24cm] at (2.63,-2.34) {$q$};
\node[cell, text width=3.15cm] at (5.95,-2.34)
  {$\{{\color{ovFit}d_1},{\color{ovFit}d_2}\}$};
\node[cell, text=ovFit, text width=2.25cm] at (9.05,-2.34)
  {\textbf{Sufficient}\\[-1pt] 1969};
\node[cell, text width=3.42cm] at (11.40,-2.34) {---};
\draw[ovInk!20, line width=.4pt] (0,-2.73) -- (14.9,-2.73);

\node[cell, text width=2.27cm] at (0.12,-3.11) {\textbf{Deletion}};
\node[cell, text width=2.24cm] at (2.63,-3.11) {$q$ \textcolor{ovInk!65}{(same)}};
\node[changed, text width=2.70cm, anchor=west] at (5.89,-3.11)
  {$\{{\color{ovFit}d_1}\}$};
\node[cell, text=ovChange, text width=2.25cm] at (9.05,-3.11)
  {\textbf{Insufficient}\\[-1pt] no birth year};
\node[cell, text width=3.42cm] at (11.40,-3.11) {Passage count: $2\to1$};
\draw[ovInk!20, line width=.4pt] (0,-3.50) -- (14.9,-3.50);

\node[cell, text width=2.27cm] at (0.12,-3.88) {\textbf{Substitution}};
\node[cell, text width=2.24cm] at (2.63,-3.88) {$q$ \textcolor{ovInk!65}{(same)}};
\node[changed, text width=2.70cm, anchor=west] at (5.89,-3.88)
  {$\{{\color{ovFit}d_1},{\color{ovChange}d_3}\}$};
\node[cell, text=ovChange, text width=2.25cm] at (9.05,-3.88)
  {\textbf{Insufficient}\\[-1pt] no birth year};
\node[cell, text width=3.42cm] at (11.40,-3.88)
  {Evidence-only patterns};
\draw[ovInk!20, line width=.4pt] (0,-4.27) -- (14.9,-4.27);

\node[cell, text width=2.27cm] at (0.12,-4.65) {\textbf{Question swap}};
\node[changed, text width=3.02cm, anchor=west, inner ysep=1pt] at (2.57,-4.65)
  {$q'$: \emph{When} $\to$ \emph{Where}};
\node[cell, text width=3.15cm] at (5.95,-4.65)
  {$\{{\color{ovFit}d_1},{\color{ovFit}d_2}\}$};
\node[cell, text=ovChange, text width=2.25cm] at (9.05,-4.65)
  {\textbf{Insufficient}\\[-1pt] no birthplace};
\node[cell, text width=3.42cm] at (11.40,-4.65)
  {Question wording};
\draw[ovInk!55, line width=.55pt] (0,-5.05) -- (14.9,-5.05);
\end{scope}
\end{tikzpicture}%
}
\caption{Paired evidence contrasts. Green marks supporting passages; orange marks the distractor, changed inputs, and insufficient outcomes. Deletion and substitution change passages with $q$ fixed; question swap changes $q$ with the passages fixed. The last column lists possible cues.}
\label{fig:overview}
\end{figure}

Evaluating evidence sufficiency requires constructing insufficient evidence, yet different constructions test different capabilities and expose different unintended cues. A score therefore reflects the construction as well as sufficiency, so we design the evaluation before the method: we construct paired sufficiency contrasts, audit which cues other than sufficiency still predict the label, and use the resulting failure regimes to motivate the three readers introduced in Section~\ref{sec:baseline}.

\subsection{Paired Sufficiency Contrasts}

Each contrast pairs a question $q$ and a sufficient passage set $E^{+}$ with an insufficient counterpart from the same underlying example. Some counterparts are rebuilt by us and others come from the source dataset. Figure~\ref{fig:overview} illustrates the three constructions:

\begin{itemize}[leftmargin=1em]
    \item \textbf{Substitution} replaces a required passage with a distractor while preserving the question and the number of passages. It is our primary construction because it holds the passage count fixed: on our HotpotQA rebuild~\citep{yang2018hotpotqa}, the passage count alone scores 0.500. Section~\ref{sec:audit} audits the cues it can leave, such as the distractor's low word overlap with the question.

    \item \textbf{Deletion} removes required support. It creates the evidence gap directly, but it also shrinks the evidence set, so a scorer could call the smaller set insufficient without reading it. We therefore take deletion pairs from NeoQA, where each question comes with many evidence sets of one to 120 passages; there, set size is only weakly tied to sufficiency, and passage count alone scores $0.563$.

    \item \textbf{Question swap} keeps the evidence fixed and pairs it with a question that is related to the passage but cannot be answered from it. This construction tests whether a scorer can tell a question the passage answers from a related one it does not. The choice of replacement question is critical. On SQuAD~2.0~\citep{rajpurkar2018squad2}, counting how many of the question's words appear in the passage separates answerable from unrelated questions almost perfectly (0.993), but scores only 0.573 against the unanswerable questions the dataset's annotators wrote about each passage. Section~\ref{sec:audit} audits the remaining question-wording cues.
\end{itemize}

\begin{table}[t]
\caption{Construction cues under shared passage additions. Columns 0, +5, +10, +15, and +20 count passages added to each side before BM25 orders the full set; the 0 condition is reordered too. Entries are pairwise accuracy (chance $0.5$; lower values point the other way). \emph{Size} (passage count) and \emph{Q-only} (question-only classifier) do not depend on $k$; minimum IDF uses the $k=0$ weights, and evidence-only is refitted at each count.}
\label{tab:cue-checks}
\begin{center}
\footnotesize
\setlength{\tabcolsep}{3pt}
\begin{tabular}{@{}lcc@{\hspace{10pt}}ccccc@{\hspace{10pt}}ccccc@{}}
\toprule
& \multicolumn{2}{c}{\textbf{Any $k$}} & \multicolumn{5}{c}{\textbf{Min. IDF coverage}} & \multicolumn{5}{c}{\textbf{Evidence-only}} \\
\cmidrule(lr){2-3}\cmidrule(lr){4-8}\cmidrule(l){9-13}
\textbf{Dataset} & Size & Q-only & $0$ & $+5$ & $+10$ & $+15$ & $+20$ & $0$ & $+5$ & $+10$ & $+15$ & $+20$ \\
\midrule
SQuAD & 0.500 & 0.607 & 0.577 & 0.524 & 0.529 & 0.536 & 0.535 & 0.500 & 0.494 & 0.472 & 0.481 & 0.484 \\
IIRC & 0.891 & 0.691 & 0.248 & 0.486 & 0.495 & 0.497 & 0.500 & 0.839 & 0.754 & 0.708 & 0.675 & 0.665 \\
NeoQA & 0.563 & 0.500 & 0.489 & 0.500 & 0.500 & 0.500 & 0.500 & 0.568 & 0.545 & 0.530 & 0.521 & 0.515 \\
HPQA & 0.500 & 0.500 & 0.941 & 0.541 & 0.520 & 0.513 & 0.510 & 0.763 & 0.758 & 0.750 & 0.745 & 0.743 \\
MuSQ & 0.501 & 0.500 & 0.524 & 0.507 & 0.504 & 0.502 & 0.502 & 0.759 & 0.763 & 0.766 & 0.767 & 0.768 \\
Wiki & 0.500 & 0.656 & 0.939 & 0.567 & 0.558 & 0.552 & 0.544 & 0.500 & 0.508 & 0.500 & 0.519 & 0.490 \\
\bottomrule
\end{tabular}
\end{center}
\end{table}

\begin{table}[t]
\caption{Cue-controlled subsets: pairwise accuracy with 95\% group-bootstrap intervals. HotpotQA passages use a fixed shuffle; each row restricts pairs to the stated control.}
\label{tab:cue-controlled-full}
\centering\footnotesize
\setlength{\tabcolsep}{3pt}
\begin{tabular*}{\linewidth}{@{\extracolsep{\fill}}llrrrrrr@{}}
\toprule
\multirow{2}{*}{\textbf{Dataset}} & \multirow{2}{*}{\textbf{Control}} & \multirow{2}{*}{\textbf{Pairs}} & \multirow{2}{*}{\textbf{\method}} & \multirow{2}{*}{\textbf{Self-RAG}} & \multirow{2}{*}{\textbf{UPR}} & \textbf{1.5B} & \textbf{API} \\
& & & & & & \textbf{yes/no} & \textbf{Suff. Ctx.} \\
\midrule
HotpotQA & min. IDF coverage tied & 771 & \textbf{0.776} & 0.665 & 0.737 & 0.619 & 0.709 \\
& & & {\scriptsize[0.75,\,0.81]} & {\scriptsize[0.65,\,0.68]} & {\scriptsize[0.71,\,0.76]} & {\scriptsize[0.58,\,0.65]} & {\scriptsize[0.69,\,0.73]} \\[1pt]
MuSiQue & answer string on both sides & 1{,}796 & \textbf{0.835} & 0.677 & 0.795 & 0.564 & 0.730 \\
& & & {\scriptsize[0.82,\,0.85]} & {\scriptsize[0.66,\,0.69]} & {\scriptsize[0.78,\,0.81]} & {\scriptsize[0.54,\,0.59]} & {\scriptsize[0.72,\,0.74]} \\[1pt]
IIRC & equal $|E|$ & 414 & 0.773 & 0.650 & 0.558 & 0.599 & \textbf{0.774} \\
& & & {\scriptsize[0.72,\,0.82]} & {\scriptsize[0.60,\,0.70]} & {\scriptsize[0.51,\,0.61]} & {\scriptsize[0.54,\,0.66]} & {\scriptsize[0.74,\,0.80]} \\[1pt]
NeoQA & equal support count & 832 & \textbf{0.644} & 0.377 & 0.438 & 0.112 & 0.567 \\
& & & {\scriptsize[0.58,\,0.71]} & {\scriptsize[0.30,\,0.46]} & {\scriptsize[0.36,\,0.52]} & {\scriptsize[0.08,\,0.15]} & {\scriptsize[0.52,\,0.61]} \\[1pt]
\bottomrule
\end{tabular*}
\end{table}

\subsection{Paired Evaluation}

We measure how often a scorer ranks the sufficient side of a contrast above the insufficient side. Records built from the same example form a group, and each sufficient record is paired with each insufficient record in its group. The share of pairs in which the sufficient side scores higher, with equal scores counting half, is the pairwise accuracy ($\wq$); 0.5 is chance and 1 is perfect ordering. It equals a within-group ROC--AUC weighted by each group's number of pairs. Confidence intervals use a cluster bootstrap with 2,000 draws that resamples complete groups, preserving dependence among records derived from the same question or passage. We report each construction separately. Pooling unrelated groups into one AUC would let between-group differences inflate the score of a scorer that fails within each example~\citep{robinson1950ecological,simpson1951interpretation}.

\subsection{Contrast Sources and Residual Cues}
\label{sec:audit}

For HotpotQA~\citep{yang2018hotpotqa}, we rebuild fixed-size contrasts by replacing an annotated support passage with a sampled distractor. MuSiQue~\citep{trivedi2022musique} supplies its own paired answerable and unanswerable variants, which we use without selecting new distractors. NeoQA~\citep{glockner2025neoqa} supplies native test deletion variants, which we use without sampling new deletions. SQuAD~2.0~\citep{rajpurkar2018squad2} and fresh Wikipedia hold the evidence fixed and vary the question, while IIRC~\citep{ferguson2020iirc} groups questions by source article, and each question keeps its own supplied evidence. Appendix~\ref{app:dataset-construction} specifies the pairing and sampling rules.

\paragraph{Residual cues.} A scorer might distinguish sufficient from insufficient records using a cue that does not establish whether the passages answer the question. We test passage count, minimum IDF coverage (the lowest share of the question's IDF-weighted words found in any single passage), and classifiers that see only the question or only the passages (Appendix~\ref{app:cue-classifiers}). Table~\ref{tab:cue-checks} shows how often each cue gives the sufficient side of a pair the higher score. Passage count alone scores $0.891$ on IIRC. A classifier reading only the question scores $0.607$--$0.691$ on SQuAD, IIRC, and fresh Wikipedia.

Adding passages hides some cues but not others. On HotpotQA, the distractor usually covers fewer of the question's important words than the passage it replaces, so minimum IDF coverage alone scores $0.941$ ($k=0$). We then add the same 5, 10, 15, or 20 passages to both sides and order each full set with BM25 (Appendix~\ref{app:padding}). With 20 additions, an added passage often has the lowest coverage on both sides. The two sides tie on this cue in $97.9\%$ of pairs, and its score falls to $0.510$. The evidence-only classifier still scores $0.743$, showing that it can use other differences in the passages.

Table~\ref{tab:cue-controlled-full} makes a second check. It keeps only pairs that match on one cue: minimum IDF coverage for HotpotQA, passage count for IIRC, number of annotated supporting passages for NeoQA, and answer-string presence for MuSiQue. \method\ has the highest score among local methods in each subset. Each check controls one cue at a time; for example, in 831 of the 832 NeoQA pairs with equal support counts, the sufficient side still has fewer passages.

\paragraph{Failure regimes.} The resulting contrasts show three distinct ways evidence can be insufficient. First, a set of passages can omit support for one part of the question, even while its remaining passages are relevant. Substitution and deletion make this \emph{missing-coverage} regime visible. Second, passages can be on topic but offer no answer-bearing span; question swaps isolate this \emph{topical-but-answerless} regime. Third, the needed facts can be spread across passages, such that no individual passage is sufficient, but their combination is. This \emph{cross-passage} regime occurs in multi-hop settings including HotpotQA, IIRC, and MuSiQue. The regimes overlap, and together they explain why the best single relevance, answerability, or generator-confidence signal differs from dataset to dataset. Section~\ref{sec:baseline} therefore builds one answer-free reader for each regime.

\section{The \method\ Scorer}
\label{sec:baseline}

Given a question $q$ and an evidence set $E$, let $h_r(q,E)$ be the raw score from reader $r\in\mathcal{R}$, where $\mathcal{R}=\{\mathrm{content},\mathrm{span},\mathrm{latent}\}$. We define \method as
\[
  s(q,E)=\sum_{r\in\mathcal{R}}\frac{h_r(q,E)-\widehat{c}_r(|E|)-\widehat{\mu}_r}{\widehat{\sigma}_r}.
\]
All hatted quantities are estimated only on the training fold. The size correction $\widehat{c}_r$ is the linear trend $\widehat{\alpha}_r+\widehat{\beta}_r\log(1+|E|)$ fitted across training records. For the span and latent readers, sets larger than any training set instead follow a size curve $\widehat{f}_r$ learned from padded training records, $\widehat{c}_r(|E|)=\widehat{\alpha}_r+\widehat{\beta}_r\log(1+N)+\widehat{f}_r(|E|)-\widehat{f}_r(N)$ for $|E|>N$, where $N$ is the largest training set. $\widehat{\mu}_r$ and $\widehat{\sigma}_r$ are the mean and standard deviation of the size-corrected scores $h_r-\widehat{c}_r$. A larger $s(q,E)$ indicates more sufficient evidence. Every component scores the input without decoding text, and the largest backbone makes one forward pass.

\begin{figure}[t]
\centering
\definecolor{reContent}{HTML}{3B6EA5}\definecolor{reSpan}{HTML}{B8672E}\definecolor{reLatent}{HTML}{7A5C99}\definecolor{reInk}{HTML}{344151}\definecolor{reFit}{HTML}{2E7D73}
\begin{tikzpicture}[
  x=1cm, y=1cm, font=\small,
  flow/.style={-{Latex[length=1.5mm]}, line width=.5pt, draw=reInk!80},
  box/.style={draw=reInk!60, fill=white, line width=.5pt, align=center, inner sep=5pt},
  offline/.style={box, dashed, fill=black!3},
  reader/.style={rounded corners=2pt, line width=.6pt, minimum width=2.75cm, minimum height=2.05cm},
  title/.style={font=\scriptsize\scshape, text=#1},
  rcap/.style={font=\tiny, text=reInk!85, text width=2.45cm, align=center},
  note/.style={align=center, font=\scriptsize\scshape, text=reInk!80},
]
\node[note, anchor=south] at (-2.5,.38) {Inference: Per Record};
\node[box, minimum width=3.8cm] (input) at (-2.5,0) {Incoming record $(q,E)$};

\node[reader, draw=reContent, fill=reContent!7] (content) at (-5.35,-1.95) {};
\node[reader, draw=reSpan,    fill=reSpan!7]    (span)    at (-2.5,-1.95)  {};
\node[reader, draw=reLatent,  fill=reLatent!7]  (latent)  at (0.35,-1.95)  {};
\node[title=reContent] at (-5.35,-1.17) {Content};
\node[title=reSpan]    at (-2.5,-1.17)  {Span};
\node[title=reLatent]  at (0.35,-1.17)  {Latent};
\node[rcap] at (-5.35,-2.68) {Is every part of the question covered?};
\node[rcap] at (-2.5,-2.68)  {Does some passage contain an answer?};
\node[rcap] at (0.35,-2.68)  {Do the passages suffice together?};

\begin{scope}[shift={(-5.35,-1.95)}]
  \foreach \i/\ok in {0/1,1/1,2/0,3/1} {
    \pgfmathsetmacro{\x}{-0.72+0.48*\i}
    \ifnum\ok=1
      \fill[reContent!45] (\x-0.17,0.26) rectangle (\x+0.17,0.44);
    \else
      \fill[red!55!black!35] (\x-0.17,0.26) rectangle (\x+0.17,0.44);
      \node[font=\tiny\bfseries, text=red!60!black] at (\x,0.12) {$\times$};
    \fi
  }
  \node[font=\tiny, text=reInk!70, anchor=east] at (-0.92,0.35) {$q$};
  \fill[black!18] (-0.9,-0.38) rectangle (-0.08,-0.26);
  \fill[black!18] (0.08,-0.38) rectangle (0.9,-0.26);
  \draw[reContent!80, line width=.4pt] (-0.72,0.26) -- (-0.5,-0.26);
  \draw[reContent!80, line width=.4pt] (-0.24,0.26) -- (-0.35,-0.26);
  \draw[reContent!80, line width=.4pt] (0.72,0.26) -- (0.55,-0.26);
\end{scope}

\begin{scope}[shift={(-2.5,-1.95)}]
  \fill[black!18] (-0.95,-0.05) rectangle (0.95,0.13);
  \fill[reSpan!75] (-0.1,-0.05) rectangle (0.48,0.13);
  \draw[reSpan, line width=.4pt] (-0.1,0.2) -- (-0.1,0.26) -- (0.48,0.26) -- (0.48,0.2);
  \node[font=\tiny, text=reSpan!90!black] at (0.19,0.39) {best span};
  \node[font=\tiny, text=reInk!70] at (0,-0.3) {vs.\ the no-answer option};
\end{scope}

\begin{scope}[shift={(0.35,-1.95)}]
  \foreach \y in {0.26,0.06,-0.14} \fill[black!18] (-1.0,\y) rectangle (-0.52,\y+0.11);
  \draw[flow, draw=reLatent] (-0.46,0.11) -- (-0.22,0.11);
  \node[circle, draw=reLatent, fill=white, inner sep=1.2pt, font=\tiny] (lm) at (0.0,0.11) {LM};
  \draw[reLatent!70, dashed, line width=.4pt] (0.3,-0.3) -- (1.0,0.46);
  \node[font=\tiny, text=reLatent] at (1.07,0.3) {$u$};
  \draw[flow, draw=reLatent] (lm.east) -- (0.62,0.05);
  \fill[reLatent] (0.64,0.07) circle (1.1pt);
\end{scope}

\draw (input.south) -- (-2.5,-.62);
\draw (-5.35,-.62) -- (0.35,-.62);
\draw[flow] (-5.35,-.62) -- (content.north);
\draw[flow] (-2.5,-.62) -- (span.north);
\draw[flow] (0.35,-.62) -- (latent.north);

\node[note, anchor=south] at (4.45,.38) {Training: Once per Fold};
\node[offline, minimum width=3.0cm] (train) at (4.45,0) {Training records $\mathcal D_{\mathrm{train}}$};
\node[reader, dashed, draw=reFit, fill=reFit!7] (fit) at (4.45,-1.95) {};
\node[title=reFit] at (4.45,-1.17) {Fold Statistics};
\node[rcap] at (4.45,-2.74) {Trend in $\log|E|$ and spread around it};
\begin{scope}[shift={(4.45,-1.95)}]
  \fill[reFit!14] (-0.85,-0.3) -- (0.9,0.24) -- (0.9,0.46) -- (-0.85,-0.08) -- cycle;
  \draw[reInk!60, line width=.35pt, -{Latex[length=1mm]}] (-1.0,-0.4) -- (1.05,-0.4);
  \draw[reInk!60, line width=.35pt, -{Latex[length=1mm]}] (-1.0,-0.4) -- (-1.0,0.52);
  \foreach \x/\y in {-0.76/-0.24,-0.64/-0.08,-0.52/-0.16,-0.4/0.0,-0.28/-0.05,-0.15/0.1,-0.03/0.02,0.1/0.18,0.22/0.1,0.35/0.28,0.47/0.18,0.6/0.35,0.72/0.26,0.84/0.4}
    \fill[reFit!70!black] (\x,\y) circle (0.8pt);
  \draw[reFit, line width=.75pt] (-0.85,-0.19) -- (0.9,0.35);
\end{scope}
\draw[flow, dashed, draw=reFit] (train.south) -- (fit.north);

\node[box, fill=reInk!4, minimum width=8.45cm, minimum height=1.55cm] (combine) at (-2.5,-4.35) {};
\node[font=\scriptsize\scshape, text=reInk] at (-2.5,-3.78) {Normalize and Combine};
\begin{scope}[shift={(-4.75,-4.42)}]
  \draw[reInk!35, line width=.3pt] (-1.75,0.0) -- (-0.35,0.0);
  \draw[reContent, line width=.5pt, fill=reContent!25] plot[domain=-1.72:-1.28, samples=30] (\x,{0.26*exp(-((\x+1.5)^2)/(2*0.07^2))});
  \draw[reSpan, line width=.5pt, fill=reSpan!25] plot[domain=-1.35:-0.75, samples=30] (\x,{0.17*exp(-((\x+1.05)^2)/(2*0.12^2))});
  \draw[reLatent, line width=.5pt, fill=reLatent!25] plot[domain=-0.78:-0.4, samples=30] (\x,{0.31*exp(-((\x+0.58)^2)/(2*0.05^2))});
  \draw[flow, draw=reInk!80] (-0.25,0.12) -- (0.12,0.12);
  \draw[reInk!35, line width=.3pt] (0.2,0.0) -- (1.5,0.0);
  \draw[reInk!45, densely dotted, line width=.35pt] (0.85,0.0) -- (0.85,0.3);
  \foreach \c/\dx in {reContent/-0.02,reSpan/0,reLatent/0.02}
    \draw[\c, line width=.55pt] plot[domain=-0.6:0.6, samples=40] (\x+0.85+\dx,{0.28*exp(-((\x)^2)/(2*0.17^2))});
  \node[font=\tiny, text=reInk!75, anchor=north] at (-1.05,-0.03) {reader scores};
  \node[font=\tiny, text=reInk!75, anchor=north] at (0.85,-0.03) {common scale};
\end{scope}
\draw[flow, draw=reInk!80] (-3.1,-4.3) -- (-2.7,-4.3);
\node[anchor=west, font=\small] at (-2.6,-4.3) {$s={\color{reContent}z_{\mathrm{content}}}+{\color{reSpan}z_{\mathrm{span}}}+{\color{reLatent}z_{\mathrm{latent}}}$};
\node[font=\tiny, text=reInk!80] at (-0.95,-4.72) {fixed equal weights};
\draw[flow, draw=reContent] (content.south) -- (content.south |- combine.north);
\draw[flow, draw=reSpan]    (span.south)    -- (span.south |- combine.north);
\draw[flow, draw=reLatent]  (latent.south)  -- (latent.south |- combine.north);
\draw[dashed, draw=reFit] (fit.south) -- (4.45,-4.35);
\draw[flow, dashed, draw=reFit] (4.45,-4.35) -- node[above, font=\scriptsize, text=reFit] {$\widehat\theta$} (combine.east);
\end{tikzpicture}
\caption{\method\ combines content coverage (blue), answer-span evidence (orange), and latent cross-passage evidence projected onto $u$ (purple). Solid paths show inference; the dashed path fits training-fold statistics for size correction and standardization before an equal-weight sum.}
\label{fig:pipeline}
\end{figure}
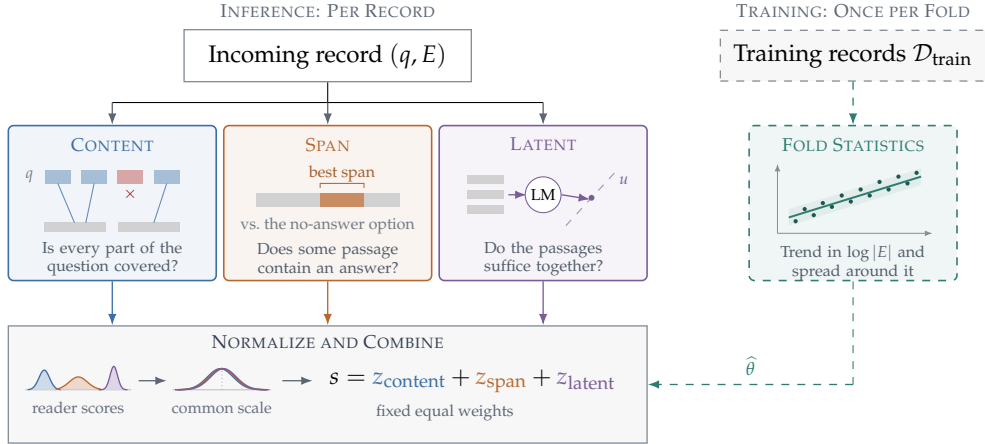

\paragraph{Content coverage.} The content reader measures whether every local part of the question is supported somewhere in the evidence. We split the question into overlapping four-WordPiece windows with stride two \citep{devlin2019bert}. Each window keeps its best match across passages under both IDF-weighted overlap \citep{sparckjones1972specificity} and a 118M multilingual reranker trained on mMARCO \citep{wang2021minilmv2,bonifacio2021mmarco}. A match must come from one source, so unrelated fragments cannot jointly satisfy a window; chunks with the same source identifier are rejoined before matching. Quantiles of the two coverage profiles feed a conditional-logistic head fitted on paired training examples \citep{breslow1978matched}.

\paragraph{Answer-span evidence.} The span reader asks whether any passage contains text that an extractive QA model prefers to its no-answer option. We train a 125M RoBERTa-base reader \citep{liu2019roberta} with explicit null supervision on Natural Questions \citep{kwiatkowski2019natural}: each negative replaces the answer sentences with a length-matched topical window (Appendix~\ref{app:span-reader}). At inference, the reader scores every passage and overflow window by its best legal span logit minus its no-answer logit, and returns the maximum margin.

\paragraph{Cross-passage evidence.} The latent reader captures evidence that becomes informative only when passages are read together. Frozen Qwen2.5-1.5B-Instruct \citep{qwen2024qwen25} encodes the question and the five highest-reranked passages in one forward pass, preserving their input order. We project the unit-normalized final-token state at layer 17 onto a fixed direction: the family-balanced mean of sufficient-minus-insufficient hidden-state differences estimated from the other training families. The five-passage cap bounds context length and exposure to irrelevant material.

\paragraph{Size correction and training.} Raw reader scores differ in scale and grow with evidence-set size; the correction and standardization above put them on one scale. Because the best reader varies across datasets (Section~\ref{sec:anatomy}), the weights stay equal and fixed rather than tuned per dataset. Training sets hold at most 22 passages while NeoQA's reach 120. To extend the correction past that range, we pad each training record with 5 to 100 passages from other questions of its dataset, which changes its size but not its label, and fit a piecewise-linear curve in $\log(1+|E|)$ with knots at the deciles within records; its slope falls from 2.9 for small sets to about 1 between 20 and 100 passages, against 3.1 for the linear fit. Only NeoQA and one IIRC record exceed the training range. On NeoQA validation timelines, a flat tail beat linear extrapolation and the padding curve matched it (Appendix~\ref{app:repro}). The padding curve and latent-reader training on the same top five passages used at inference together raise NeoQA from $0.673$ to $0.773$. The neural backbones are frozen and shared across datasets. \method\ outputs a ranking score; each deployment thresholds it to answer, abstain or retrieve more.

\section{Experiments}
\label{sec:autorater}

\paragraph{Evaluation protocol.} We compare scorers on the same sufficient--insufficient pairs from six datasets using pairwise accuracy (Section~\ref{sec:protocol}); the cue checks use these records too. HotpotQA substitution passages are shuffled once, so the distractor's position carries no signal (Appendix~\ref{app:dataset-construction}). We assemble 13{,}600 training records from MuSiQue \citep{trivedi2022musique}, 2WikiMultiHopQA \citep{ho2020constructing}, IIRC \citep{ferguson2020iirc}, SQuAD~2.0 \citep{rajpurkar2018squad2}, HotpotQA \citep{yang2018hotpotqa}, and CRAG \citep{yang2024crag}. For each evaluated family, we fit the content head, latent direction, and normalization without records from that family. The span reader is trained once on Natural Questions~\citep{kwiatkowski2019natural} and shared across datasets. We compare with methods based on passage relevance \citep{asai2024selfrag,sachan2022upr,es2024ragas}, retrieval control \citep{cheng2024uar,yan2024crag,jiang2023flare}, and confidence in a generated answer \citep{kadavath2022ptrue,honovich2022true,gao2023alce,kuhn2023semantic,farquhar2024semantic}. The closest baseline, Sufficient Context \citep{joren2025sufficient}, asks a model directly whether the context is sufficient; we run its official prompt on GPT-6 Luna and on Qwen2.5-1.5B-Instruct \citep{qwen2024qwen25}. Other local methods use the same 1.5B model where possible, and the UAR, CRAG, and Self-RAG classifiers keep their own backbones (Appendix~\ref{app:methods}).

\begin{table}[t]
\caption{Pairwise accuracy (chance $0.5$), single-record latency (ms), and saturated throughput (records/s). Datasets: SQuAD~2.0, IIRC, NeoQA, HotpotQA substitution (HPQA), MuSiQue (MuSQ), and fresh Wikipedia (Wiki). Avg/Min summarize the six datasets; timings use 57 records from each on one GPU. Shading marks the best three local methods; brackets give 95\% group-bootstrap intervals (Avg combines per-dataset errors). HPQA passages use a fixed shuffle.}
\label{tab:grid}
\begin{center}
\scriptsize
\setlength{\tabcolsep}{2.5pt}
\begin{tabular}{@{}llrrrrrr!{\hspace{5pt}}rr!{\hspace{5pt}}|rrr@{}}
\toprule
\textbf{Method} & \textbf{Size} & \textbf{SQuAD} & \textbf{IIRC} & \textbf{NeoQA} & \textbf{HPQA} & \textbf{MuSQ} & \textbf{Wiki} & \textbf{Avg} & \textbf{Min} & \textbf{p50} & \textbf{p95} & \textbf{rec/s} \\
& & \scriptsize Q-swap & \scriptsize Native & \scriptsize Delete & \scriptsize Subst. & \scriptsize Subst. & \scriptsize Q-swap & & & \multicolumn{2}{c}{\scriptsize ms} & \\
\cmidrule(r){1-2}\cmidrule(lr){3-8}\cmidrule(lr){9-10}\cmidrule(l){11-13}
\multicolumn{13}{@{}l}{\textit{Frontier model through an API (reference)}} \\
Suff.\ Context & API & 0.799 & 0.813 & 0.676 & 0.764 & 0.788 & 0.864 & 0.784 & 0.676 & \multicolumn{3}{c}{API call} \\
 & & {\tiny[0.79,\,0.81]} & {\tiny[0.80,\,0.83]} & {\tiny[0.66,\,0.69]} & {\tiny[0.76,\,0.77]} & {\tiny[0.78,\,0.79]} & {\tiny[0.85,\,0.87]} & {\tiny[0.78,\,0.79]} & & \multicolumn{3}{c}{} \\[1pt]
\midrule
\multicolumn{13}{@{}l}{\textit{Local, reads the question and evidence}} \\
Suff.\ Context & 1.5B & 0.539 & 0.584 & 0.514 & 0.521 & 0.514 & 0.567 & 0.540 & 0.514 & 5566 & 5934 & 0.43 \\
 & & {\tiny[0.53,\,0.55]} & {\tiny[0.57,\,0.60]} & {\tiny[0.51,\,0.52]} & {\tiny[0.51,\,0.53]} & {\tiny[0.51,\,0.52]} & {\tiny[0.55,\,0.58]} & {\tiny[0.53,\,0.54]} & & & & \\[1pt]
yes/no prompt & 1.5B & \silver{0.720} & 0.579 & 0.500 & 0.684 & 0.569 & 0.933 & 0.664 & 0.500 & \silver{31.8} & \gold{424} & 8.1 \\
 & & {\tiny[0.70,\,0.74]} & {\tiny[0.56,\,0.60]} & {\tiny[0.49,\,0.51]} & {\tiny[0.67,\,0.70]} & {\tiny[0.56,\,0.58]} & {\tiny[0.92,\,0.94]} & {\tiny[0.66,\,0.67]} & & & & \\[1pt]
UPR & 1.5B & \bronze{0.699} & 0.541 & \silver{0.699} & \silver{0.712} & \bronze{0.731} & \silver{0.946} & \bronze{0.721} & 0.541 & \gold{24.5} & \silver{537} & \bronze{9.8} \\
 & & {\tiny[0.68,\,0.72]} & {\tiny[0.52,\,0.56]} & {\tiny[0.69,\,0.71]} & {\tiny[0.70,\,0.72]} & {\tiny[0.72,\,0.74]} & {\tiny[0.93,\,0.96]} & {\tiny[0.72,\,0.73]} & & & & \\[1pt]
RAGAS-CR & 1.5B & 0.588 & 0.485 & 0.508 & 0.549 & 0.513 & 0.665 & 0.551 & 0.485 & 77.9 & 1764 & 5.1 \\
 & & {\tiny[0.57,\,0.60]} & {\tiny[0.48,\,0.49]} & {\tiny[0.50,\,0.51]} & {\tiny[0.54,\,0.56]} & {\tiny[0.51,\,0.52]} & {\tiny[0.65,\,0.68]} & {\tiny[0.55,\,0.56]} & & & & \\[1pt]
CRAG & 770M & 0.626 & 0.584 & 0.549 & 0.609 & 0.633 & 0.794 & 0.632 & \bronze{0.549} & 39.4 & 968 & 5.7 \\
 & & {\tiny[0.61,\,0.65]} & {\tiny[0.56,\,0.61]} & {\tiny[0.54,\,0.56]} & {\tiny[0.60,\,0.62]} & {\tiny[0.63,\,0.64]} & {\tiny[0.77,\,0.81]} & {\tiny[0.63,\,0.64]} & & & & \\[1pt]
UAR-tree & 7B & 0.527 & 0.467 & 0.481 & 0.500 & 0.502 & 0.506 & 0.497 & 0.467 & 72.6 & 780 & 3.1 \\
 & & {\tiny[0.52,\,0.54]} & {\tiny[0.45,\,0.48]} & {\tiny[0.48,\,0.48]} & {\tiny[0.50,\,0.50]} & {\tiny[0.50,\,0.50]} & {\tiny[0.49,\,0.52]} & {\tiny[0.49,\,0.50]} & & & & \\[1pt]
Self-RAG & 7B & \gold{0.752} & \silver{0.688} & \bronze{0.644} & \bronze{0.696} & \silver{0.759} & \bronze{0.939} & \silver{0.746} & \silver{0.644} & 44.3 & 2041 & 2.8 \\
 & & {\tiny[0.73,\,0.77]} & {\tiny[0.67,\,0.71]} & {\tiny[0.63,\,0.65]} & {\tiny[0.69,\,0.70]} & {\tiny[0.75,\,0.77]} & {\tiny[0.93,\,0.95]} & {\tiny[0.74,\,0.75]} & & & & \\[1pt]
\midrule
\multicolumn{13}{@{}l}{\textit{Local, reads a generated answer}} \\
P(True) & 1.5B & 0.670 & 0.607 & 0.520 & 0.590 & 0.571 & 0.854 & 0.635 & 0.520 & 164 & \bronze{587} & 7.4 \\
 & & {\tiny[0.65,\,0.69]} & {\tiny[0.58,\,0.63]} & {\tiny[0.51,\,0.53]} & {\tiny[0.58,\,0.60]} & {\tiny[0.56,\,0.58]} & {\tiny[0.84,\,0.87]} & {\tiny[0.63,\,0.64]} & & & & \\[1pt]
Attrib. & 1.5B & \bronze{0.699} & 0.615 & 0.517 & 0.570 & 0.551 & 0.929 & 0.647 & 0.517 & 142 & 630 & 9.3 \\
 & & {\tiny[0.68,\,0.72]} & {\tiny[0.59,\,0.64]} & {\tiny[0.51,\,0.52]} & {\tiny[0.56,\,0.58]} & {\tiny[0.54,\,0.56]} & {\tiny[0.92,\,0.94]} & {\tiny[0.64,\,0.65]} & & & & \\[1pt]
SemEnt & 1.5B & 0.593 & 0.602 & 0.505 & 0.539 & 0.544 & 0.589 & 0.562 & 0.505 & 499 & 1382 & 2.4 \\
 & & {\tiny[0.57,\,0.61]} & {\tiny[0.58,\,0.62]} & {\tiny[0.50,\,0.51]} & {\tiny[0.53,\,0.55]} & {\tiny[0.53,\,0.55]} & {\tiny[0.57,\,0.61]} & {\tiny[0.56,\,0.57]} & & & & \\[1pt]
FLARE & 1.5B & 0.586 & \bronze{0.627} & 0.503 & 0.539 & 0.565 & 0.666 & 0.581 & 0.503 & 184 & 778 & \silver{11.9} \\
 & & {\tiny[0.57,\,0.61]} & {\tiny[0.60,\,0.65]} & {\tiny[0.49,\,0.51]} & {\tiny[0.53,\,0.55]} & {\tiny[0.55,\,0.58]} & {\tiny[0.64,\,0.69]} & {\tiny[0.57,\,0.59]} & & & & \\[1pt]
\midrule
\textbf{\method} & 1.5B & 0.684 & \gold{0.926} & \gold{0.773} & \gold{0.853} & \gold{0.813} & \gold{0.973} & \gold{\textbf{0.837}} & \gold{\textbf{0.684}} & \bronze{36.5} & 603 & \gold{17.4} \\
 & & {\tiny[0.66,\,0.70]} & {\tiny[0.91,\,0.94]} & {\tiny[0.76,\,0.78]} & {\tiny[0.84,\,0.86]} & {\tiny[0.80,\,0.82]} & {\tiny[0.97,\,0.98]} & {\tiny[0.83,\,0.84]} & & & & \\[1pt]
\bottomrule\end{tabular}
\end{center}
\end{table}

\paragraph{Main results.} \method\ has both the highest average ($0.837$, against $0.746$ for Self-RAG, the strongest prior method) and the highest minimum (Table~\ref{tab:grid}). Asked directly whether the evidence suffices, the Sufficient Context prompt scores $0.540$ on the local 1.5B model and $0.784$ on the frontier model, whose yes-or-no labels tie on $26$--$57\%$ of pairs, each counted as one half under $\wq$. \method\ leads the frontier model on five of six datasets, with a mean paired difference of $+0.053$ \ci{+0.047}{+0.059}. \method's weakest dataset, SQuAD~2.0, still scores $0.684$; every other local method drops to $0.644$ or lower somewhere, and eight of the eleven drop to $0.520$ or lower. \method\ leads on IIRC, NeoQA, HotpotQA, MuSiQue and fresh Wikipedia, spanning all three constructions and IIRC's native contrast. Among local methods, Self-RAG leads on SQuAD~2.0 ($0.752$ versus $0.684$ for \method). On NeoQA, \method's content reader alone reaches $0.799$, above the full sum ($0.773$) and every other local scorer (Section~\ref{sec:anatomy}). NeoQA's news articles still contain plausible answer spans after the needed fact is deleted, so the span reader scores near chance there ($0.508$) and lowers the sum.

Methods that read a generated answer fail on NeoQA: all four score $0.503$--$0.520$. NeoQA's events are invented, so the generator cannot have memorized them, yet it stays confident in answers its evidence does not support. UAR's decision tree scores $0.497$ on average, at chance.

\paragraph{Real retrieval output.} To test \method\ beyond constructed contrasts, we compare the top five passages returned by BM25 and e5-base-v2 for each question. Across five datasets, we retain 3{,}149 pairs for which one retrieved set contains all annotated supporting passages and the other does not (Table~\ref{tab:real-retrieval}). With the evaluated family held out during fitting, \method\ averages $0.813$, ahead of Self-RAG ($0.674$), UPR ($0.661$) and the yes/no prompt ($0.590$), and leads the local methods on four of the five datasets; UPR leads on SQuAD~2.0 ($0.871$ vs.\ $0.813$). Both sets hold five passages, so set size carries no signal, and always choosing one retriever's set scores $0.204$--$0.796$; \method\ beats both choices on every dataset. The frontier model averages $0.800$ on the same pairs, below \method, and leads on 2WikiMultiHopQA ($0.899$ vs.\ $0.840$).

\begin{table}[t]
\caption{Pairwise accuracy on 3{,}149 BM25-versus-dense retrieval pairs; 2Wiki is 2WikiMultiHopQA and the other abbreviations follow Table~\ref{tab:grid}. The last two rows always choose the named retriever's set. \emph{Avg}: unweighted mean over datasets. \textbf{Bold}: best local method.}
\label{tab:real-retrieval}
\begin{center}\footnotesize
\setlength{\tabcolsep}{3.2pt}
\begin{tabular}{@{}lrrrrr!{\hspace{6pt}}r@{}}
\toprule
& \textbf{SQuAD} & \textbf{IIRC} & \textbf{MuSQ} & \textbf{HPQA} & \textbf{2Wiki} & \textbf{Avg} \\
Pairs & 481 & 556 & 608 & 1{,}147 & 357 & \\
\midrule
Suff.\ Context (API) & 0.810 & 0.725 & 0.785 & 0.783 & 0.899 & 0.800 \\
\midrule
\method & 0.813 & \textbf{0.766} & \textbf{0.808} & \textbf{0.837} & \textbf{0.840} & \textbf{0.813} \\
UPR & \textbf{0.871} & 0.583 & 0.664 & 0.588 & 0.601 & 0.661 \\
Self-RAG & 0.798 & 0.589 & 0.677 & 0.678 & 0.627 & 0.674 \\
yes/no prompt & 0.636 & 0.662 & 0.572 & 0.616 & 0.462 & 0.590 \\
\midrule
dense side & 0.536 & 0.426 & 0.340 & 0.796 & 0.742 & 0.568 \\
BM25 side & 0.464 & 0.574 & 0.660 & 0.204 & 0.258 & 0.432 \\
\bottomrule
\end{tabular}
\end{center}
\end{table}

\paragraph{Overlap and added passages.} We remove exact matches to evaluation questions and passages from the span reader's Natural Questions training data, then audit residual shared sentences; overlap is highest in HotpotQA and MuSiQue (Appendix~\ref{app:span-reader}). Adding the same 5--20 passages to both sides of each group and BM25-ordering the full sets lowers \method's mean $\wq$ from $0.847$ at $k=0$ to $0.804$ at $k=20$ (Appendix~\ref{app:padding}).

\paragraph{Cost.} On one NVIDIA A100 GPU, \method\ takes 36.5~ms per record at the median, versus 24.5~ms for UPR and 31.8~ms for the 1.5B yes/no prompt. It has the highest local throughput, 17.4 records/s versus FLARE's 11.9. Its 603~ms 95th-percentile latency comes mostly from the 118M reranker (465~ms); the 1.5B pass takes 29.0~ms (Table~\ref{tab:grid}; Appendix~\ref{app:repro}).

Because \method\ runs before the generator, it can also decide whether to call it. As a gate in front of the 12 generators of Table~\ref{tab:motivation}, thresholded on training folds and applied to the same 600 records, it lowers the mean rate of answering on insufficient evidence from $0.636$ to $0.511$ and saves $13.3\%$ of generation time, while the rate of answering sufficient records falls from $0.861$ to $0.783$.

\section{Component Analysis}
\label{sec:anatomy}

\begin{table}[!htb]
\caption{Reader ablation, pairwise accuracy. Every subset uses the same size correction and equal-weight sum. Abbreviations follow Table~\ref{tab:grid}; \emph{Avg} and \emph{Min} are over the six datasets. \textbf{Bold}: best per column.}
\label{tab:branch-ablation}
\begin{center}\footnotesize
\setlength{\tabcolsep}{3.2pt}
\begin{tabular}{@{}lrrrrrr|rr@{}}
\toprule
\textbf{Readers} & \textbf{SQuAD} & \textbf{IIRC} & \textbf{NeoQA} & \textbf{HPQA} & \textbf{MuSQ} & \textbf{Wiki} & \textbf{Avg} & \textbf{Min} \\
\midrule
\method\ (all three) & 0.684 & \textbf{0.926} & 0.773 & 0.853 & \textbf{0.813} & \textbf{0.973} & \textbf{0.837} & \textbf{0.684} \\
\midrule
content + span & 0.618 & 0.912 & 0.740 & \textbf{0.887} & 0.785 & 0.958 & 0.817 & 0.618 \\
content + latent & 0.664 & 0.772 & 0.764 & 0.835 & 0.796 & 0.962 & 0.799 & 0.664 \\
span + latent & \textbf{0.716} & 0.925 & 0.607 & 0.756 & 0.731 & 0.940 & 0.779 & 0.607 \\
content & 0.579 & 0.607 & \textbf{0.799} & 0.861 & 0.762 & 0.941 & 0.758 & 0.579 \\
span & 0.635 & 0.904 & 0.508 & 0.644 & 0.625 & 0.855 & 0.695 & 0.508 \\
latent & 0.712 & 0.827 & 0.583 & 0.714 & 0.702 & 0.927 & 0.744 & 0.583 \\
\bottomrule
\end{tabular}
\end{center}
\end{table}

Table~\ref{tab:branch-ablation} evaluates every subset of the three readers.
We observe that each reader leads where its failure regime dominates. For example, content coverage is the strongest single reader on the evidence-removal contrasts, NeoQA ($0.799$), HotpotQA ($0.861$) and MuSiQue ($0.762$), where the remaining passages stay on topic and still hold answer-like spans. The span reader leads on IIRC ($0.904$), and the latent reader on SQuAD~2.0 ($0.712$), whose authored near-misses reuse the passage's vocabulary.
A deployed scorer cannot tell which contrast a record comes from, so \method\ sums all three readers. Removing any reader lowers the average: to $0.817$ without latent, $0.799$ without span, and $0.779$ without content. The best single reader's weakest dataset scores $0.583$, against $0.684$ for the sum. The full sum has the best average and minimum of all seven subsets.

\section{Conclusion}

Retrieved passages can be relevant without being sufficient, and generators rarely notice the gap. Instead of relying on the generator, we asked whether sufficiency can be judged from the question and evidence alone, before generation.
Testing this requires care, because direct ways of constructing insufficient evidence can leave shortcuts that solve the test without judging sufficiency; we therefore built paired contrasts and audited their residual cues. The contrasts show three ways evidence falls short: the question's content is missing, on-topic passages offer no answer, or support is spread across passages. No single reader covers all three, so we propose \method, which sums three complementary readers, one per failure. In our evaluation, \method\ has the highest average and the highest weakest-dataset score of all compared methods, including a frontier model asked the same question. The advantage carries over to real retriever output, and \method\ takes only 36.5~ms per question on one GPU. Sufficiency can thus be judged cheaply and locally, before any generator is called.


\subsection*{Acknowledgments}

\subsection*{Statement on the Use of AI}
Generative AI tools were used during the preparation of this manuscript to assist with language and formatting, literature retrieval and discovery, research ideation and execution, including artifact implementation and experimental support, and the generation of synthetic questions for the fresh Wikipedia evaluation set described in Appendix~\ref{app:dataset-construction}. The authors reviewed the resulting text, equations, tables, figures, and citations, and take responsibility for the final content.

\subsection*{Ethics Statement}
This work uses publicly available QA datasets and Wikipedia content and involves no human subjects or collection of private data. The synthetic questions may inherit biases from the generating model; their construction and filtering are documented in Appendix~\ref{app:dataset-construction}. 


\bibliography{references}
\bibliographystyle{colm2026_conference}

\appendix
\raggedbottom
\clearpage
\section{Adaptation of Comparison Methods}
\label{app:methods}

We adapt the comparison methods in Table~\ref{tab:grid} as follows. Passage-level scores use the maximum over evidence items; answer-aware methods start from one greedy answer to the full set, with 32{,}768-token prompts that are not truncated. Adapted local prompts use Qwen2.5-1.5B-Instruct \citep{qwen2024qwen25}, entailment uses mDeBERTa-v3-base (279M; \citealp{he2023debertav3}), and released classifiers retain their own weights.

\paragraph{Sufficient Context (frontier and local).} \citet{joren2025sufficient} ask whether a context contains enough information to answer. We use their published autorater instructions on local Qwen2.5-1.5B-Instruct \citep{qwen2024qwen25} and on GPT-6 Luna through Amazon Bedrock's \texttt{us.openai.gpt-6-luna} inference profile \citep{aws2026gpt6luna}. For the Table~\ref{tab:grid} frontier row, we fill the released prompt's timestamp with 2026-09-02, cap each passage at 1{,}600 characters, and make one API call for each of the 74{,}841 records. We set a 4{,}096-token completion limit and leave other inference options at their defaults. The parsed JSON sufficiency label is a binary score, so two records with equal labels count as a tie under $\wq$ ($26$--$57\%$ of pairs across datasets); two responses without a parseable label receive 0.

\paragraph{Yes/no prompt.} This control uses the same local Qwen2.5-1.5B-Instruct, but replaces the published Sufficient Context prompt with the short template below and scores token probabilities instead of a generated JSON label. It lists every evidence item in its evaluation input order as \texttt{[1]}, \texttt{[2]}, \ldots, cutting each to 800 characters. The text is one user turn in the model's chat template, with the assistant turn opened; the full prompt is not truncated and no answer is decoded. Its score is the last-position log-odds $\log\sum_{t\in Y}e^{\ell_t}-\log\sum_{t\in N}e^{\ell_t}$, where $Y$ and $N$ contain first tokens of \emph{yes} and \emph{no} in upper and lower case, with and without a leading space.
\begin{quote}\small\ttfamily\raggedright
Judge whether the documents below contain enough information for a diligent reader to derive a definitive answer to the question. Combining facts across documents is allowed; guessing beyond the documents is not.\\[3pt]
Question: \{q\}\\[3pt]
Documents:\\
\{docs\}\\[3pt]
Do the documents contain enough information to answer the question? Answer with one word, yes or no.
\end{quote}

\paragraph{UPR.} \citet{sachan2022upr} rerank passages by the likelihood of the query given each passage. We use their prompt with the local 1.5B model, take mean $\log P(q\mid p)$ for each passage, and keep the maximum.

\paragraph{RAGAS-CR.} The reference-free context-relevance measure of \citet{es2024ragas} extracts sentences needed to answer. We use its official prompt and score the fraction of context sentences extracted.

\paragraph{CRAG.} The retrieval evaluator of \citet{yan2024crag} grades a passage as correct, incorrect, or ambiguous. We apply its released T5-large classifier (770M; \citealp{raffel2020t5}) to \texttt{query [SEP] passage} and take the maximum correct-class logit.

\paragraph{UAR-tree.} \citet{cheng2024uar} use a four-head decision tree to decide whether an instruction needs retrieval. We use the released Llama-2-7B backbone \citep{touvron2023llama2} and the official DROP template \citep{dua2019drop}, reading every passage in its evaluation input order; the score is $1$ for ``no retrieval''. The 4{,}096-token limit prefix-truncates 13{,}080 of 17{,}572 NeoQA records and 309 of 16{,}000 MuSiQue records, and none elsewhere.

\paragraph{Self-RAG.} \citet{asai2024selfrag} use reflection tokens during generation. We use only the released 7B critic, without decoding, and take the maximum passage-wise difference $\operatorname{logit}(\texttt{[Relevant]})-\operatorname{logit}(\texttt{[Irrelevant]})$.

\paragraph{P(True).} \citet{kadavath2022ptrue} judge whether a proposed answer is correct. Given the shared greedy answer, we read the local model's last-position probability of \emph{True} against \emph{False}.

\paragraph{Attrib.} Attribution checks whether a source supports a generated claim \citep{honovich2022true,gao2023alce}. We turn the shared answer $a$ into ``the answer to $q$ is $a$'' and take the maximum passage-level entailment score.

\paragraph{SemEnt.} Semantic entropy groups answers by meaning to measure uncertainty \citep{kuhn2023semantic,farquhar2024semantic}. We cluster the greedy answer and five samples by bidirectional entailment, then use negative cluster entropy.

\paragraph{FLARE.} \citet{jiang2023flare} trigger retrieval when a forthcoming sentence has low probability. We score the shared greedy answer by its minimum token log-probability, without running another retrieval step.

\section{Evaluation Dataset Construction}
\label{app:dataset-construction}

The six evaluation families in Table~\ref{tab:grid} use the following group and sampling rules. Pairwise accuracy includes every sufficient--insufficient pair in a group containing both labels.

\paragraph{HotpotQA substitution.} We start with HotpotQA~\citep{yang2018hotpotqa} records that have at least two annotated supporting passages, at least four passages in the original set, and a nonempty question. A pseudorandom generator with seed 17 selects one supporting passage to replace. One distractor is drawn from a different record; a second is drawn from a different record whose passage title shares a token longer than three characters with the removed passage's title. Both insufficient sets retain the other original passages and thus have the same size as the sufficient set. We keep a group only when both distractors can be found, yielding 4{,}117 groups and 8{,}234 pairs. Passage order is then shuffled with a fixed seed for evaluation.

\paragraph{MuSiQue substitution.} We use the answerable and unanswerable variants supplied by MuSiQue-Full~\citep{trivedi2022musique}; we do not select replacement passages. The dataset authors assemble a 20-passage context from supporting passages and retrieved distractors, using component questions with intermediate answers hidden for retrieval. For an unanswerable variant, they choose a constituent hop and require its answer to be absent from every context passage. We pair variants sharing the dataset's group identifier and question.

\paragraph{NeoQA deletion.} We use the native test variants from the main, optimal-evidence, and context-ablation subsets~\citep{glockner2025neoqa}, retaining their \emph{answerable-sufficient} and \emph{answerable-insufficient} labels and excluding native unanswerable questions. We group variants by timeline and question family; no new test deletion is sampled. In optimal-evidence variants, an insufficient set omits a required article from a minimal sufficient coalition. In the main variants, it removes the articles carrying a selected required outline fact, which may be more than one article. Context-ablation variants add nested same-timeline nonsupporting articles around two required articles and omit each required article in turn. We compare every supplied sufficient and insufficient variant within a mixed-label group.

\paragraph{SQuAD~2.0 and IIRC.} SQuAD~2.0~\citep{rajpurkar2018squad2} groups the dataset's authored answerable and unanswerable questions by source paragraph, keeping that paragraph fixed. IIRC~\citep{ferguson2020iirc} groups records by source article and retains the 1{,}327 groups with both native answerability labels. Every answerable question is paired with every unanswerable question in its group, yielding 3{,}653 pairs. Each IIRC record keeps its own supplied evidence set, so both question and evidence may change within a pair.

\paragraph{Fresh Wikipedia.} We use passages from English Wikipedia articles created after July 2026. For each group, \texttt{gpt-oss-120b} writes one answerable question and up to two related unanswerable questions about a fixed passage. Unanswerable edits include replacing an entity or adding a premise whose key entity is absent; string and alias checks require the positive answer span to appear and the substituted or premise entity to be absent. We pool 1{,}294 mechanically verified candidate groups, then run three rounds of question-text-only filtering, removing the most predictable 8\% of groups in each round. The retained set has 1{,}006 groups, 2{,}918 records, and 1{,}912 pairs.

\section{Construction-Cue Measures}
\label{app:cue-classifiers}

\paragraph{Passage count.} We score each record by the number of passages in its evidence set, $|E|$.

\paragraph{Minimum IDF coverage.} For each passage, we compute the share of the question's IDF weight carried by words in that passage and take the minimum across the evidence set. In 7{,}966 one-for-one HotpotQA substitutions, the inserted distractor covers a median $4.6\%$ of that weight, versus $44.8\%$ for the support passage it replaces. The distractor is the least-covered passage of the insufficient set in $94.0\%$ of these pairs. Shared added passages can make the two sets' minima equal while the original replacement remains.

\paragraph{Question-only classifier.} We predict sufficiency from the question text alone. This classifier and the evidence-only classifier use word unigrams and bigrams hashed into 16{,}384 dimensions with a four-byte BLAKE2b digest. Features are weighted by $\log(1+\mathrm{tf})$ times training-fold IDF, $\log((n+1)/(\mathrm{df}+1))+1$, and normalized to unit length. We fit an L2-regularized logistic regression with an intercept using 300 full-batch gradient steps (learning rate 0.5, L2 coefficient 1). We assign complete question groups to five folds using seed 17 and score each record with the model trained on the other four folds. Models are fit separately for each dataset.

\paragraph{Evidence-only classifier.} We apply the same feature map and training procedure to all passages joined with blank lines in their BM25 order. For the passage-addition analysis, we refit this classifier at each $k\in\{0,5,10,15,20\}$ after BM25 ordering. The out-of-fold sufficiency probabilities from both classifiers are evaluated with the pairwise accuracy of Section~\ref{sec:protocol}.

\section{RINSE Training and Runtime}
\label{app:repro}

Train and test question groups are disjoint. Backbones and preprocessing are shared across datasets; each fold fits the content head, latent direction, and normalization statistics.

\paragraph{\method.} The \method\ row of Table~\ref{tab:grid} sums the three size-corrected reader scores with equal weights (Section~\ref{sec:baseline}); it reads the question and evidence without generating an answer.

\paragraph{Size correction beyond the training range.} Training sets hold at most 22 passages, while NeoQA reaches 120. On held-out validation timelines (3{,}159 records from 117 question groups), a flat tail raised $\wq$ from $0.664$ to $0.730$ over linear extrapolation. The padding curve, specified before test evaluation, scored $0.729$, within the preset $0.01$ tolerance. We therefore use the linear correction within the training range and the learned curve beyond it. Training the latent reader on the same top five passages it reads at inference then raised validation $\wq$ to $0.757$; together the changes raised NeoQA test performance from $0.673$ to $0.773$ (paired difference $+0.100$ \ci{+0.091}{+0.108}).

\paragraph{Timing.} We sample 57 records from each of the six Table~\ref{tab:grid} datasets and time single-record latency on one GPU with one resident model copy. For throughput, independent warmed replicas process the same 342 records concurrently; we double the replica count from one to 16 and report the first trial reaching 90\% mean GPU utilization or 90\% peak memory use, dividing completed records by the slowest replica's wall-clock time.

\section{Span-Reader Training and Overlap Audit}
\label{app:span-reader}

We train pretrained RoBERTa-base once on Natural Questions \citep{kwiatkowski2019natural}, without earlier QA fine-tuning. Each positive uses an answer-bearing passage; its negative replaces answer sentences with length-matched topical text, or uses an answer-free passage if replacement is impossible. We discard negatives in which a gold answer string survives, and exact matches to evaluation questions or passages.

The data contain 90{,}732 training and 16{,}013 validation pairs. We train for one epoch with AdamW (batch 32, learning rate $2\times10^{-5}$, 6\% warm-up, weight decay 0.01, bfloat16), 384-token inputs, and spans of at most 30 tokens; this takes 12 minutes on one GPU. The score is the best span logit sum minus the no-answer logit sum at \texttt{CLS}, maximized over windows and passages. External validation reaches $0.994$ pairwise accuracy and $0.995$ ROC--AUC, above the preset thresholds of $0.80$ and $0.85$.

\paragraph{Passage-overlap audit.} We remove exact question and source-passage matches, then check for shared sentences of at least eight words. Passage overlap is absent in NeoQA and fresh Wikipedia, low in SQuAD~2.0 and IIRC, and reaches $29.5\%$ of HotpotQA and $97.8\%$ of MuSiQue question groups ($26.6\%$ of MuSiQue passages). These are passage matches, not question matches.

\section{RINSE as Evidence Sets Grow}
\label{app:padding}

Adding passages from other questions weakens \method's pairwise accuracy even when both sides receive the same new material. At $k\in\{0,5,10,15,20\}$, we give every record in a question group the same first $k$ added passages, order the complete set by BM25 without filtering, and apply the unchanged scorer. Across six datasets, pairwise accuracy falls monotonically from $0.847$ to $0.804$ ($\Delta=-0.043$ \ci{-0.048}{-0.039}). BM25 also reorders the original passages and recomputes reader features at $k=0$, so Table~\ref{tab:padding} can differ from Table~\ref{tab:grid} before any passages are added.

The loss is uneven. IIRC drops from $0.925$ to $0.790$ ($\Delta=-0.135$), and NeoQA from $0.777$ to $0.708$ ($\Delta=-0.069$) despite starting with a median of 41 passages. SQuAD~2.0 and fresh Wikipedia lose $0.033$ and $0.019$, while HotpotQA and MuSiQue change by less than $0.005$. On IIRC the max-over-passages span reader falls from $0.904$ to $0.686$; on fresh Wikipedia the latent reader falls from $0.927$ to $0.849$. On NeoQA the combined score loses more pairwise accuracy than any reader alone.

The added passages keep their original labels. A screen for answer strings, repeated titles and same-question sufficient passages flags at most 8.3\% of insufficient records in paired groups at $k=20$. Excluding every flagged pair changes no estimate in Table~\ref{tab:padding} by more than $0.002$.

\begin{table}[H]
\caption{Sensitivity to shared passage additions. Values are pairwise accuracy ($\wq$) with 95\% group-bootstrap intervals; $\Delta$ compares $k=20$ with $k=0$.}
\label{tab:padding}
\centering\small
\setlength{\tabcolsep}{4pt}
\begin{tabular}{@{}lrrrrrrl@{}}
\toprule
\textbf{Dataset} & \textbf{Pairs} & $k{=}0$ & $k{=}5$ & $k{=}10$ & $k{=}15$ & $k{=}20$ & $\boldsymbol{\Delta\wq}$, $k{=}20$ \\
\midrule
SQuAD 2.0 & 2{,}803 & 0.684 & 0.666 & 0.661 & 0.655 & 0.651 & $-0.033$ \ci{-0.049}{-0.019} \\
 & & {\tiny[0.66,\,0.70]} & {\tiny[0.65,\,0.69]} & {\tiny[0.64,\,0.68]} & {\tiny[0.63,\,0.67]} & {\tiny[0.63,\,0.67]} & \\[1pt]
IIRC & 3{,}653 & 0.925 & 0.860 & 0.823 & 0.801 & 0.790 & $-0.135$ \ci{-0.152}{-0.118} \\
 & & {\tiny[0.91,\,0.94]} & {\tiny[0.84,\,0.88]} & {\tiny[0.81,\,0.84]} & {\tiny[0.78,\,0.82]} & {\tiny[0.77,\,0.81]} & \\[1pt]
NeoQA & 102{,}876 & 0.777 & 0.745 & 0.731 & 0.718 & 0.708 & $-0.069$ \ci{-0.076}{-0.062} \\
 & & {\tiny[0.77,\,0.79]} & {\tiny[0.73,\,0.76]} & {\tiny[0.72,\,0.74]} & {\tiny[0.71,\,0.73]} & {\tiny[0.70,\,0.72]} & \\[1pt]
HotpotQA & 8{,}234 & 0.896 & 0.897 & 0.895 & 0.893 & 0.893 & $-0.003$ \ci{-0.007}{+0.002} \\
 & & {\tiny[0.89,\,0.90]} & {\tiny[0.89,\,0.91]} & {\tiny[0.89,\,0.90]} & {\tiny[0.88,\,0.90]} & {\tiny[0.88,\,0.90]} & \\[1pt]
MuSiQue & 8{,}000 & 0.828 & 0.828 & 0.828 & 0.829 & 0.827 & $-0.001$ \ci{-0.006}{+0.004} \\
 & & {\tiny[0.82,\,0.84]} & {\tiny[0.82,\,0.84]} & {\tiny[0.82,\,0.84]} & {\tiny[0.82,\,0.84]} & {\tiny[0.82,\,0.83]} & \\[1pt]
fresh Wikipedia & 1{,}912 & 0.973 & 0.970 & 0.963 & 0.962 & 0.954 & $-0.019$ \ci{-0.028}{-0.010} \\
 & & {\tiny[0.97,\,0.98]} & {\tiny[0.96,\,0.98]} & {\tiny[0.95,\,0.97]} & {\tiny[0.95,\,0.97]} & {\tiny[0.94,\,0.96]} & \\[1pt]
\midrule
Average & & 0.847 & 0.828 & 0.817 & 0.810 & 0.804 & $-0.043$ \ci{-0.048}{-0.039} \\
 & & {\tiny[0.84,\,0.85]} & {\tiny[0.82,\,0.83]} & {\tiny[0.81,\,0.82]} & {\tiny[0.80,\,0.82]} & {\tiny[0.80,\,0.81]} & \\
\bottomrule
\end{tabular}
\end{table}

\section{Extended Related Work}
\label{app:related}

A RAG system can respond to insufficient evidence at multiple points in its pipeline. Before generation, sufficiency can be assessed from the question and retrieved evidence alone; this is the decision made by \method. Relevance scoring operates at the same stage but asks a different question. During or after generation, retrieval controllers, confidence estimators, and verification methods instead rely on the generator’s internal states or outputs. Finally, abstention can be trained directly into the generator.

\paragraph{Evidence sufficiency.} Sufficient Context defines the same target as \method---whether a diligent reader could answer from the provided context alone---and labels it with an autorater ranging from an 11B entailment model to a frontier LLM, reporting that many instances whose context is annotated as gold are in fact insufficient \citep{joren2025sufficient}. S2G-RAG judges sufficiency together with the remaining gap inside an iterative retrieval loop \citep{li2026s2g}. \method builds on these foundations and addresses two complementary questions: how negative construction and residual cues affect the validity of sufficiency evaluation, and whether sufficiency can be estimated accurately, locally, and at substantially lower cost than a large autorater. 

\paragraph{Relevance estimation and passage grading.} Relevance is a closely related and more extensively studied problem that asks whether an individual passage is pertinent to a question. 
UPR ranks by the likelihood of the query given the passage \citep{sachan2022upr}, RAGAS elicits a graded judgment from a model \citep{es2024ragas}, and CRAG grades each retrieved document to decide whether the retrieval must be corrected \citep{yan2024crag}. Relevance differs from sufficiency in two respects: a passage can be about the question and still omit the fact that the answer requires, and a per-document grade cannot capture cases in which the evidence becomes sufficient only when multiple passages are considered together.

\paragraph{Generator-side signals.} Unlike sufficiency and relevance scorers that operate on the question and evidence alone, another line of work derives its signal from the answer generator’s internal states or outputs. 
Among methods based on internal states, UAR classifies a frozen model's hidden state to decide whether retrieval is needed at all \citep{cheng2024uar}, while Probing-RAG probes hidden states to predict answer correctness \citep{baek2025probingrag}; related work separates what a model knows from what retrieval supplies \citep{zhou2026ralm,julka2026lockin}. These studies motivate \method’s latent reader (Section \ref{sec:baseline}), although their prediction targets are tied to a particular generator. 

Other methods rely on generated outputs. P(True) asks a model how much it trusts its own answer \citep{kadavath2022ptrue}; semantic entropy compares sampled answers \citep{kuhn2023semantic,farquhar2024semantic}; and attribution methods test whether the evidence entails a generated claim \citep{honovich2022true,gao2023alce}. Adaptive-retrieval controllers act earlier but on the same information: FLARE retrieves again when an upcoming token is improbable \citep{jiang2023flare}, Self-RAG trains reflection tokens into the generator \citep{asai2024selfrag}, and later systems predict whether another round will help or whether an answer has become stable \citep{park2025stoprag,qiu2026sure,yang2025simrag,kieback2026tasr,moskvoretskii2025adaptive,tian2026predicting,zhang2026utility,setiawan2026abstention}. In contrast, \method targets whether the supplied evidence is sufficient independently of the downstream generator.

\paragraph{Training abstention into the generator.}
R-Tuning fine-tunes it to say \emph{I don't know} on questions it cannot answer \citep{zhang2024rtuning}, Trust-Align aligns it for grounded refusal and citation \citep{song2025trustalign}, and TruthRL and GRACE optimize rewards that separate correct answers from hallucinations and abstentions, GRACE adding an explicit sufficiency term to the reward \citep{wei2025truthrl,zhao2026grace}. The resulting behavior belongs to one model and does not survive further training: reinforcement finetuning for reasoning reduces refusal on unanswerable questions \citep{song2025hallucinationtax}, and reasoning fine-tuning lowers abstention across twenty datasets \citep{kirichenko2025abstentionbench}. A generator-independent score is complementary to this line, since it supplies the sufficiency signal such rewards must otherwise construct, and it outlives any particular generator.

\paragraph{Evaluation with constructed negatives.} Work on constructed negatives asks whether they test the intended capability. Natural language inference labels are partly predictable from the hypothesis alone \citep{gururangan2018artifacts,poliak2018hypothesis}, models exploit lexical overlap in place of entailment \citep{mccoy2019right}, and SQuAD~2.0 replaced automatically generated unanswerable questions with authored near-misses for precisely this reason \citep{rajpurkar2018squad2}. We apply the same discipline to evidence sufficiency: Section~\ref{sec:protocol} specifies the contrast construction and audits what remains predictable from cues other than sufficiency.

\end{document}